\documentclass[runningheads]{llncs}
\usepackage[T1]{fontenc}
\usepackage{graphicx}
\usepackage{multirow}
\usepackage{subfig}
\usepackage{tabularx}
\usepackage{xcolor}

\begin{document}

\title{MAP-Elites with Transverse Assessment for Multimodal Problems in Creative Domains}
\titlerunning{MEliTA for Multimodal Problems in Creative Domains}
\author{
Marvin Zammit\orcidID{0000-0002-5343-0077}
\and
Antonios Liapis\orcidID{0000-0001-5554-1961}
\and
Georgios N. Yannakakis\orcidID{0000-0001-7793-1450}
}
\authorrunning{M. Zammit et al.}
\institute{Institute of Digital Games, University of Malta, Msida, Malta
\\
\email{\{marvin.zammit, antonios.liapis, georgios.yannakakis\}@um.edu.mt}}

\maketitle

\begin{abstract}
The recent advances in language-based generative models have paved the way for the orchestration of multiple generators of different artefact types (text, image, audio, etc.) into one system. Presently, many open-source pre-trained models combine text with other modalities, thus enabling shared vector embeddings to be compared across different generators. Within this context we propose a novel approach to handle multimodal creative tasks using Quality Diversity evolution. Our contribution is a variation of the MAP-Elites algorithm, MAP-Elites with Transverse Assessment (MEliTA), which is tailored for multimodal creative tasks and leverages deep learned models that assess coherence across modalities. MEliTA decouples the artefacts' modalities and promotes cross-pollination between elites. As a test bed for this algorithm, we generate text descriptions and cover images for a hypothetical video game and assign each artefact a unique modality-specific behavioural characteristic. Results indicate that MEliTA can improve text-to-image mappings within the solution space, compared to a baseline MAP-Elites algorithm that strictly treats each image-text pair as one solution. Our approach represents a significant step forward in multimodal bottom-up orchestration and lays the groundwork for more complex systems coordinating multimodal creative agents in the future.

\keywords{MAP-Elites \and Quality Diversity \and Image Generation \and Text Generation \and Text-to-image Generation \and Digital Games.}
\end{abstract}

\section{Introduction}
\label{sec:introduction}

Evolutionary search in creative domains such as visual, audio, or text generation has traditionally struggled to evaluate the artefacts it produces. This is mostly because there is no universal metric to assess the quality of media content \cite{ritchie2007empirical,Galanter2019problems}. Early approaches relied on ad-hoc metrics such as timing intervals in music generation \cite{alfonseca2007geneticalgomusic} and compression-based indices for image generation \cite{machado2015complexity}, or tasked humans to evaluate the evolving population \cite{secretan2008picbreeder,liapis2012adaptivemodel,takagi2001interactive,hoover2011interactively}. As more refined deep learning algorithms became available, models trained for a specific task have been employed more frequently as a fitness measure of evolved artefacts \cite{johnson2019guidance,roziere2021EvolGAN}. 

For generative media, the most interesting development in the field of deep learning is the training and release of multimodal models. These models map multiple modalities to the same latent space, thereby enabling the direct comparison of different types of media. Contrastive Language-Image Pre-Training (CLIP) \cite{openai2021clip} was one such model which demonstrated excellent zero-shot image classification to any input set of semantic labels. Similar models map text with other modalities, such as audio \cite{copet2023simple}, which in unison with CLIP (or similar models) may compare images to another modality via an intermediary text modality. Alternatively, models such as Meta's ImageBind \cite{girdhar2023imagebind} directly combine multiple input and output modalities into a single embedding space, which facilitates \textit{multimodal generation} and assessment but also opens up new possibilities for cross-modal learning and transfer learning.

The advent of more nuanced metrics based on large-scale corpora (unimodal or multimodal) does not quite address the limitations of optimising a universal ``quality'' metric in creative domains. Such a singular drive may lead to a narrow view of human creativity which often builds on niches such as art movements, music genres, and literary paradigms. To address this, more recent research in evolutionary computation has focused on the diversity of the output instead of its quality \cite{lehman2011novelty}, or combining the two in Quality Diversity (QD) algorithms \cite{pugh2016quality,gravina2019pcgqd}. QD algorithms promote diversity in the artefacts while maintaining some minimal criteria on quality \cite{lehman2010mcns,liapis2015ecj} or keeping only the fittest individuals within each phenotypic niche \cite{mouret2015illuminating}. The latter approach is followed in Novelty Search with Local Competition \cite{lehman2011novelty}, which only compares neighbouring individuals to assess their (local) dominance. Another prominent QD algorithm is Multi-dimensional Archive of Phenotypic Elites (MAP-Elites) \cite{mouret2015illuminating}. MAP-Elites partitions the solution space into a multi-dimensional grid (the \emph{feature map}), where each axis represents varying properties within a specific behavioural characteristic (BC) or phenotypic trait of the solutions. Each cell stores the optimal individual (elite) according to the global fitness function, promoting only competition within the phenotypic niche. The most popular implementation of MAP-Elites operates in a steady-state fashion, selecting a parent among the elites (at random) and mutating it to produce an offspring. The offspring is then mapped to a cell of the feature map according to its BCs and may replace the elite in that cell if it has a higher fitness. As MAP-Elites illuminates a problem space, it is particularly apt for creative domains where it has already shown successes \cite{fontaine2021differentiable,gravina2019pcgqd,viana2022enemies,alvarez2019dungeon,alvarez2022tropetwist,colton2021styletransfer}.

This paper applies the MAP-Elites algorithm to a multimodal creative domain, specifically generating text descriptions and cover images for hypothetical video games. To address this challenge, we propose an algorithmic improvement on QD search: MAP-Elites with Transverse Assessment (MEliTA). MEliTA introduces an inter-modal evaluation process that shares partial artefacts (e.g. image or text) among phenotypically similar elites in order to find more coherent pairings. This innovative approach enhances the creative co-evolutionary process, resulting in the discovery of fitter and more diverse outcomes.

\section{MAP-Elites with Transverse Assessment}
\label{sec:methodology}

MAP-Elites with Transverse Assessment (MEliTA) is a variant of MAP-Elites designed to evolve multimodal artefacts to minimise incongruity between modalities. MEliTA builds on a number of assumptions, which in our use case revolve around two modalities but can scale to any number of modalities:
\begin{itemize}
    \item each evolved individual ($A$) is a collections of $N$ (separable) artefacts, each encompassing a single modality $M_i$, e.g. $A=\{a_{M1},a_{M2},\ldots\}$
    \item variation operators ($V_{M1}$,$V_{M2},\ldots$) can be applied on each artefact type ($a_{M1},a_{M2},\ldots$) separately, potentially informed by other modalities but not modifying the other artefact types, e.g. $a_{M1}'=V_{M1}(a_{M1},a_{M2},\ldots)$ etc.
    \item there exists a function ($f$) that returns a value $q$ indicating coherence between all modalities, e.g. $q(A)=f(a_{M1},a_{M2},\ldots)$ 
    \item there exist $N$ functions ($g_{M1},g_{M2},\ldots$) each returning a value ($\beta_{M1},\beta_{M2},\ldots$) indicating properties of one artefact type ($a_{M1},a_{M2},\ldots$) separately, e.g. $\beta_{M1}(A)=g_{M1}(a_{M1})$, $\beta_{M2}(A)=g_{M2}(a_{M2})$, etc.
\end{itemize}

Following the above notations, MEliTA produces an $N$-dimensional archive of elites ($N$ being the number of modalities in the artefacts), characterised by $N$ behaviour characterisations ($\beta_{M1},\beta_{M2},\ldots$). 

During the evolutionary cycle, an existing elite $E=\{e_{M1},e_{M2},\ldots\}$ is selected from the archive---the selection operator can be uniform selection or more sophisticated \cite{cully2017qd}.
One artefact of this individual is chosen randomly (e.g. $e_{M1}$) and changed via the appropriate variation operator, creating in this example $e_{M1}'=V_{M1}(e_{M1},e_{M2},\ldots)$. 
The new artefact ($e_{M1}'$) is assigned a behaviour characterisation based on its modality (i.e. $g_{M1}(e_{M1}')$).
A new individual $E'$ is created by combining the new artefact with unchanged artefacts of the parent $E$, i.e. $E'=\{e'_{M1},e_{M2},\ldots\}$.
In vanilla MAP-Elites (as applied in this paper), the individual $E'$ would be compared with the elite ($E'_{old}$) with BCs $g_{M1}(e_{M1}'), g_{M2}(e_{M2}),\ldots$ in terms of $q$, replacing it if $q(E')>q(E'_{old})$ or occupying the cell if no elite exists for those BCs.
In MEliTA, the artefact $e_{M1}'$ is iteratively paired with the artefacts of other modalities of each elite $R=\{r_{M1},r_{M2},\ldots\}$ that occupies a cell with BC $g_{M1}(e_{M1}')$, producing a new candidate solution $R'=\{e_{M1}',r_{M2},\ldots\}$ and computing the coherence between modalities for the new individual ($q(R')$).
The collection of candidate solutions $\vec{R'}$ along with $E'$ are sorted by their coherence score $q$ and form an ordered list of candidate solutions $\vec{L}$. In order, each member $\phi\in\vec{L}$ is checked against the occupying elite ($\phi_{old}$) in a cell with the BCs of $\phi$. If no such elite exists (the cell is empty), the individual $\phi$ occupies that cell and the process ends. If the current elite $\phi_{old}$ is worse than the new candidate ($q(\phi_{old})<q(\phi)$) then $\phi$ replaces $\phi_{old}$ and the process ends; if the current elite is not worse, then the process continues with the next member in $\vec{L}$. 
This results in only one new individual being inserted into the archive (at most) per evolutionary cycle, and ensures that empty cells can also be filled in the archive (by individual $E'$, if it is better than other alternatives).
The process of MEliTA will become clearer through the use case of Section~\ref{sec:usecase}.

\section{Use Case: Generating Text \& Visuals for Game Titles}
\label{sec:usecase}

As an exploratory use case, we select images and text as modalities for MEliTA, as both benefit from the availability of reliable AI generators. In this use case, the goal is to generate fictitious video games in the form of a \textit{game title}, a \textit{short description} of the game, and a \textit{cover image}. Through this experiment, we wish to create a diverse set of coherent and appropriately game-like art and text blurbs that can inspire players and game developers alike. 

For each modality, core considerations are how the artefact is generated (or changed via mutation) and how it is characterised for the purposes of the MAP-Elites feature map \cite{mouret2015illuminating}. The sections below clarify how artefacts of each modality are generated and characterised, followed by a rundown of the MEliTA process.

\subsection{Text Modality}\label{sec:usecase_text_generation}
The Generative Pre-trained Transformer 2 (GPT-2) \cite{radford2019language} is an auto-regressive model based on the transformer architecture \cite{vaswani2017attention}. This model has undergone extensive pre-training through a substantial corpus of English text through a self-supervised learning approach \cite{balestriero2023cookbook}. 
While GPT-2 has since been eclipsed by more cogent models \cite{brown20gpt3}, particularly the more recent large-language models (LLMs) \cite{openai2023gpt4,touvron2023llama,touvron2023llama2}, it distinguishes itself with considerably faster inference, at the expense of reduced performance. For an iterative evolutionary algorithm, the substantial speed gain GPT-2 offers was considered a good trade-off.

\subsubsection{Text Generation.} 

To generate believable titles and descriptions for fictitious games, a GPT-2 language model was fine-tuned on a dataset composed of real titles and descriptions from the extensive catalogue of the Steam platform\footnote{https://store.steampowered.com/}. The data was curated, removing entries without English text, non-game entries (e.g. utilities, videos), and entries labelled with adult material or nudity. The resulting dataset contains approximately 72,000 pairs of game titles and descriptions.

For game titles, the pre-trained GPT-2 model of \cite{radford2019language} was fine-tuned exclusively on video game titles of the above Steam dataset. Given the modest scale of this dataset, we used the most compact variant of the model (approximately 124 million parameters). The transformer was trained on the list of titles demarcated by distinctive beginning and end tokens: ``\texttt{<\textbar{begin}\textbar>game title<\textbar{end}\textbar>}''. First attempts exhibited over-fitting, and the model echoed existing titles. To mitigate this, the weights and biases of the last layer were reset prior to the fine-tuning process. This resulted in a more robust model capable of generating novel titles exhibiting minimal overlap with their original counterparts.

A second GPT-2 model was fine-tuned from the pre-trained model of \cite{radford2019language}, this time incorporating both title and description (see example in Table \ref{tab:desc_mutation}) in the format ``\texttt{<\textbar{begin}\textbar>game title<\textbar{body}\textbar>description<\textbar{end}\textbar>}''. A new game title can thus be generated through the first model using an input prompt of ``\texttt{<\textbar{begin}\textbar>}''. A description can be generated by priming the second model with this new title in the format ``\texttt{<\textbar{begin}\textbar>game title<\textbar{body}\textbar>}''.

\subsubsection{Text Mutation.}
Two variation operators may be applied during mutation of text descriptions: one (partial mutation) retains coherence with the previous description while the other (full mutation) resets the description in order to avoid early convergence. For partial mutation, the first part of the description was retained, split along a selected space or punctuation mark around the middle of the text\footnote{We randomly select three spaces or punctuation marks within the text and keep the middle one. This makes it likely that the split will be in the middle of the description.}. The rest of the description is removed. The game title together with the first description fragment are reintroduced as input to the second GPT-2 model to complete the description. For full mutation, the entire description is removed and only the game title is used as input to the second GPT-2 model to produce a new description from scratch. The chance of full mutation is 20\%, otherwise partial mutation is applied. Text variation samples are shown in Table~\ref{tab:desc_mutation}.

\begin{table}[t]
\small
    \centering
    \caption{Text variation samples: Partial mutation chooses a space or punctuation at random, removes the text after it and uses the sequence up to that point to generate the remaining description (in red). Full mutation removes the description and only uses the game title (in bold) to generate the description (in red).}
    \label{tab:desc_mutation}
    \begin{tabular}{|p{0.11\textwidth}|p{0.87\textwidth}|}
        \hline
        Original & \tiny{\textbf{Hooey! You Got a Monster!?,} You've been selected for an experiment at the University of Chicago's Animal Research Center. A girl, who has lost her memory in this creepy and magical place will use your memories to discover why you got kidnapped by Professor Teller on his way there...}\\
        \hline
        Partial mutation & \tiny{\textbf{Hooey! You Got a Monster!?,} You've been selected for an experiment at the University of Chicago's Animal Research Center. \textcolor{red}{A professor with no memories about his past finds you. An unforgettable story about two animals, their lives and what happened to them in this dream-like place.}}\\
        \hline
        Full mutation & \tiny{\textbf{Hooey! You Got a Monster!?,} \textcolor{red}{The best game of all time is back, and better than ever. This year we've reimagined the classic arcade platformer with over 100 levels to conquer. Gameplay has been revamped for even more fun, it's more difficult then ever but much harder in this new version!}}\\
        \hline
    \end{tabular}
\end{table}

\subsubsection{Text Characterisation.}\label{sec:usecase_text_bc}
Several approaches to characterising text were explored for this use case, e.g. via the Gunning-Fog readability index \cite{gunning1973writing}. We settled on topic modelling using topics extracted from the Steam dataset, as we want to match against dominant patterns for game descriptions, typical \cite{ritchie2007empirical} of the domain we attempt to emulate. We used the Latent Dirichlet Allocation \cite{blei2003latentdirichlet} algorithm (LDA) for topic modelling and trained the algorithm across different numbers of topics (from 4 to 30). Using the perplexity and complexity metrics as guides, we settled on an optimal setup of 16 topics. 

To characterise each generated description, we subject it to the tuned LDA topic model, which results in a set of probabilities designating the likelihood of the description to be aligned with each of the predefined topics. In situations where the most probable topic assignment falls short of achieving a probability threshold of at least 40\% above other topics, the description is deemed as \textit{unclassified} and is not added to the feature map (i.e. it is ignored during evolution).

\subsection{Image Modality}\label{sec:usecase_image_generation}
We leverage Stable Diffusion (SD) for the image generation tasks in this paper as it can produce high-quality images with low compute. SD \cite{rombach2022latentdiffusion} has openly accessible source code and pre-trained weights. Where diffusion models \cite{sohldickstein2015deep} are trained to reconstruct noisy versions of the desired output (e.g. image), SD models  are trained on a noisy latent vector of the output, thus reducing inference time and improving robustness. Using a classifier-free guidance approach \cite{ho2021classifierfree}, both a conditional and an unconditional diffusion model are trained, and by comparing their responses during inference, a balance can be struck between image quality and faithful image adherence to the input prompt.

\subsubsection{Image Generation.}
For this paper we utilise a text-to-image SD model to generate the cover art for the fictitious games. The input textual prompts contain both a hypothetical game title and a corresponding description, presented in the format ``\texttt{title, description}'' (see example in Table \ref{tab:desc_mutation}). Based on early experiments, we also add a set of negative prompts (``duplication, ugly, text, bad anatomy'') to enhance the aesthetic quality of the generated outputs.

\begin{figure}[t]
\centering
    \includegraphics[width=0.95\textwidth]{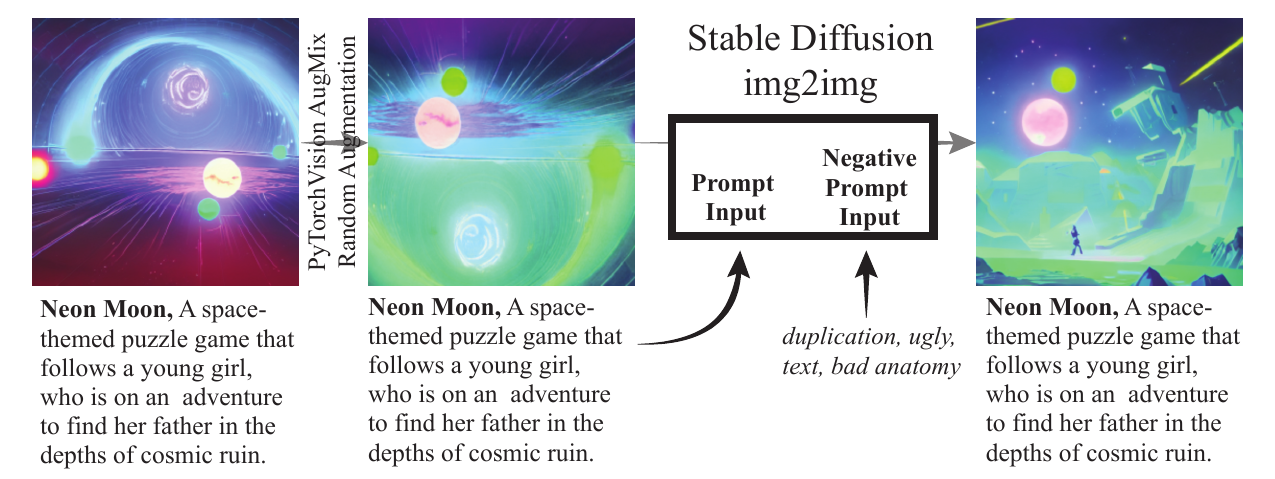}
    \caption{Image variation sample: The parent's (unchanged) text modality is used as a prompt for image repair based on SD, alongside ``standard'' negative prompts.}
    \label{fig:img_mutation}
\end{figure}

\subsubsection{Image Mutation.}
As with text, we mutate the image on the phenotype level in two stages. First, we apply the AugMix augmentation function \cite{hendrycks2020augmix} from the TorchVision software library \cite{marcel2010torchvision} to distort the original image. The distorted image is then paired with the original prompts (game title, description, negative prompts) as inputs to an image-to-image SD model (see Fig.~\ref{fig:img_mutation}).
In order to strike a balance between image quality and computational efficiency, all tasks pertaining to image generation and mutation were executed over a sequence of 40 diffusion steps. Since the augmentation function may lead to more or less distorted images, resulting images after SD may match the parent image to a lesser or greater degree. Unlike text mutation, we have less control over the chances of a large phenotypic change, but consider such a change beneficial to avoid early convergence and a slow evolutionary process.

\subsubsection{Image characterisation.}\label{sec:usecase_visual_bc}
As with text classification we explored several BCs for images, but we settled on two straightforward metrics: complexity and colourfulness. 
\textit{Image complexity} is calculated as the ratio of edge pixels found via the Holistically-Nested Edge Detection (HED) model \cite{xie15hed} over the total pixel count. While previous approaches in evolutionary art relied on Sobel or Canny filters \cite{machado2015complexity} for complexity estimation, HED edges offer a notable advantage in terms of accuracy and noise reduction.
\textit{Image colourfulness} is calculated via a quantitative measure of the perceived chromatic richness or saturation \cite{hasler2003colourfulness}. To combine the two image metrics into a concise BC, each numerical value (complexity, colourfulness) is categorised into four bins (very low, low, high, very high) and the image is classified in terms of the combinations of these bins. Therefore, the image BC comprises a total of 16 bins, matching the dimensions of the text BC.

\subsection{MEliTA applied to the use case}\label{sec:usecase_melita}

\begin{figure}[t]
    \centering
    \includegraphics[width=0.99\textwidth]{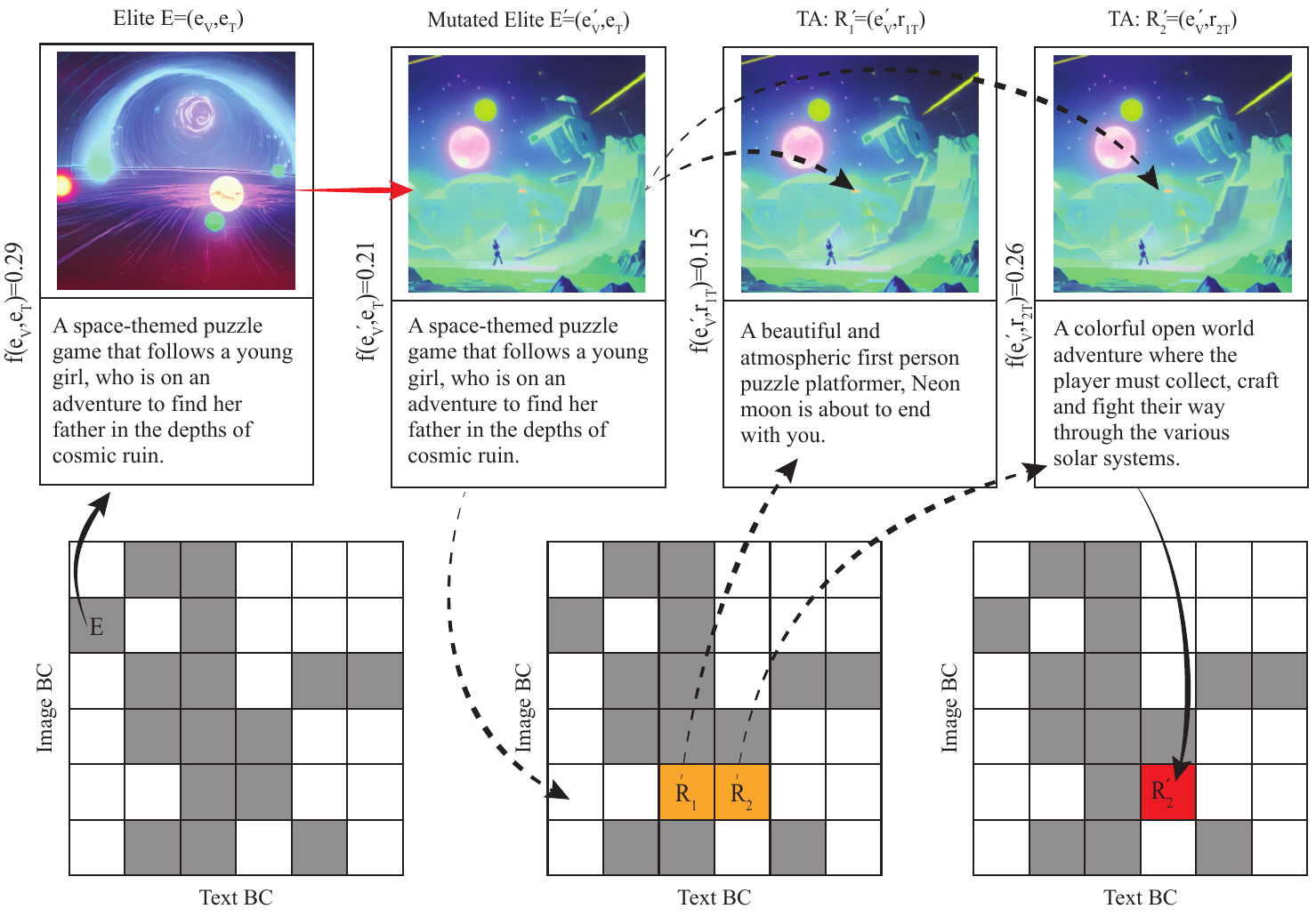}
    \caption{The MEliTA process in a simplified feature map for this use case, with grey cells occupied by elites. From one selected elite E, the changed image ($e'_V$) produces three candidate solutions from elites $E$, $R_1$, $R_2$. Based on their CLIP score, the ordered list of candidates is $\vec{L}=\{R'_2,E',R'_1\}$. Since $q(R'_2)>q(R_2)$ the candidate $R'_2$ (that merges the image from $E'$ and text from $R_2$) replaces $R_2$. If $q(R'_2){\leq}q(R_2)$ then $E'$ would occupy the empty cell at (5,0). Dotted lines denote temporary individuals that are lost after this parent selection.}
    \label{fig:melita_TA}
\end{figure}

MEliTA in this bimodal use case performs parent selection via an Upper Confidence Bound (UCB) algorithm \cite{kocsis2006montecarlo}, which takes into account frequency of parent selection and can improve coverage of the final elites \cite{sfikas2021montecarloelites}. The feature map consists of a grid of $16\times16$ cells, with a text BC for membership in one of 16 topics found in the Steam database (see Section \ref{sec:usecase_text_bc}) and a visual BC for membership in one of 16 combinations of image complexity and colourfulness categories (see Section \ref{sec:usecase_visual_bc}). This use-case uses the CLIP score between image and description (including the title as shown in Table \ref{tab:desc_mutation}) as fitness, to decide whether elites survive in the archive. In `vanilla' MAP-Elites the selected parent produces either a new image or a new description via variation operators, then pairs it with the remaining artefact (description or image, respectively) in its genotype to produce a new offspring. However, Transverse Assessment operates as follows.

When a parent is selected in MEliTA, either a new image or a new description (chosen randomly) is produced from the parent's existing artefacts via variation operators described in Sections \ref{sec:usecase_text_generation} and \ref{sec:usecase_image_generation}. Figure~\ref{fig:melita_TA} shows an example where the image ($e_V$) is modified (to $e'_V$), and will be described below. The new image $e'_V$ is paired with the parent's existing text ($e_T$) and the CLIP score, text BC and visual BC coordinates of this new candidate solution $E'$ are calculated (in this case, placing it at the currently empty cell at 5,0). For transverse assessment the image is also paired with the title and description of all elites with the same image BC of $E'$ (the row with $R_1$, $R_2$ in Fig.~\ref{fig:melita_TA}), creating new candidate solutions; their CLIP score is calculated\footnote{The BC coordinates for these candidate solutions do not need to be recalculated as they are combinations of text BCs and visual BCs that are already known.}. Starting from the fittest of these temporary solutions, if the CLIP score of an existing elite in the archive with these coordinates is lower than the new solution or if the cell is unoccupied, the candidate solution occupies that cell and the process stops (in this case replacing $R_2$). If no candidate solution is added to the archive, the process ends and a new parent selection is made.

\section{Experimental Protocol}
\label{sec:protocol}

In order to evaluate the performance of MEliTA as a QD algorithm, we leverage the bimodal generation challenge described in Section \ref{sec:usecase} and aim to validate the high-level hypothesis that \emph{MAP-Elites with Transverse Assessment leads to better quality and diversity in the generated artefacts than MAP-Elites without Transverse Assessment}. 
We formalise this into more concise hypotheses:

\begin{itemize}
\item[H1] The final archive of solutions is better and more diverse for MEliTA compared to MAP-Elites without transverse assessment.
\item[H2] MEliTA discovers better and more diverse solutions faster than MAP-Elites without transverse assessment.
\item[H3] The final archive of solutions for MEliTA can be perceived as more diverse by humans compared to MAP-Elites without transverse assessment.
\end{itemize}
H1 assesses the final product of evolution, while H2 assesses the performance of the evolutionary process over time. Both H1 and H2 assess diversity based on the ad-hoc BCs of the feature map, while H3 uses orthogonal diversity metrics which are aligned to human perceptions of diversity per modality. 

\subsubsection{Performance Metrics.}

To evaluate H1 and H2, we rely on traditional performance metrics for QD evolutionary search \cite{mouret2015illuminating}. 
Specifically, for H1, we evaluate the final archive's mean fitness (among occupying elites), max fitness (i.e. the highest fitness among elites), coverage (i.e. ratio of occupied cells over all cells in the grid) and QD score (i.e. sum of all the elites' fitness). The first two metrics measure quality, coverage measures diversity (according to the chosen BCs) and QD score measures a combination of the two\footnote{Unlike \cite{mouret2015illuminating}, we do not normalise the values to the maximum found across runs and across methods. Instead, we present the non-normalised results (e.g. the ratio of occupied versus the maximum size of the feature map for coverage).}. 
For H2, the same metrics are tracked over time (i.e. after every parent selection) and assessed as an area under the curve (AUC); high AUC values may mean that the final metrics have high scores or that reaching a high score was done earlier in the process. 
For H3, the final elites are assessed in terms of orthogonal measures of diversity to those used to populate the feature map. For visual diversity two measures are used: the Learned Perceptual Image Patch Similarity (LPIPS) measure is used as a distance metric trained on human annotations of visual distance \cite{zhang18lpips}, and the structural similarity (SSIM) measure commonly used for quantifying image degradation suffered through transmission or compression losses. For textual diversity in the descriptions, their embeddings generated by the SBERT encoder \cite{reimers2019sentencebert} are employed, since they capture the semantics of each sentence. Latent vectors of two individuals' descriptions are compared in terms of cosine similarity \cite{dangeti2017Statistics} 
to derive the \emph{SBERT distance}. For each distance metric, we calculate for each elite the mean distance with all other elites, as well as the nearest-neighbour distance with the closest elite as a more reserved measure of visual/text overlap.

\subsubsection{Test Cases.}
In order to verify these hypotheses, we follow the below protocol. 
We generated 100 game titles via the GPT-2 model described in Section \ref{sec:usecase_text_generation} and select 7 titles which are varied in terms of theme and length. These titles are:
\begin{itemize}
\item[T1] ``Neon Moon''
\item[T2] ``Lion King''
\item[T3] ``Hexgrave''
\item[T4] ``Fantasy Fables: The Legend of the Flying Sword''
\item[T5] ``The Princess of Thieves''
\item[T6] ``The Shadow Warrior 2: Shadows of the Past''
\item[T7] ``Hooey! You Got a Monster!?''	
\end{itemize}
For each title, we perform 10 evolutionary runs per tested method (MEliTA and MAP-Elites). To provide a fair and controllable initial population for these methods, we use GPT-2 to produce 100 descriptions per title and for each description we produce 4 images via SD. In each evolutionary run, all text descriptions and a random image among the 4 candidates per description produce the initial 100 individuals which are then assigned to the feature map.

\section{Results}
\label{sec:results}

In order to validate the hypotheses of Section \ref{sec:protocol}, MEliTA and MAP-Elites with no Transverse Assessment were ran with the same initial populations per game title (minor variation introduced through stochastic image selection as described in Section \ref{sec:protocol}) for 2000 parent selections. Results throughout this section referring to the final archive are derived from the elites in the feature map after 2000 parent selections. Statistical significance between the results of 10 evolutionary runs of different methods is established via the non-parametric two-tail Wilcoxon Rank-Sum test, at a significance level $p<0.05$.

\subsection{Evaluating the quality and diversity in the final archive}\label{sec:results_h1}

\begin{figure}[t]
    \centering
    \includegraphics[width=0.98\textwidth]{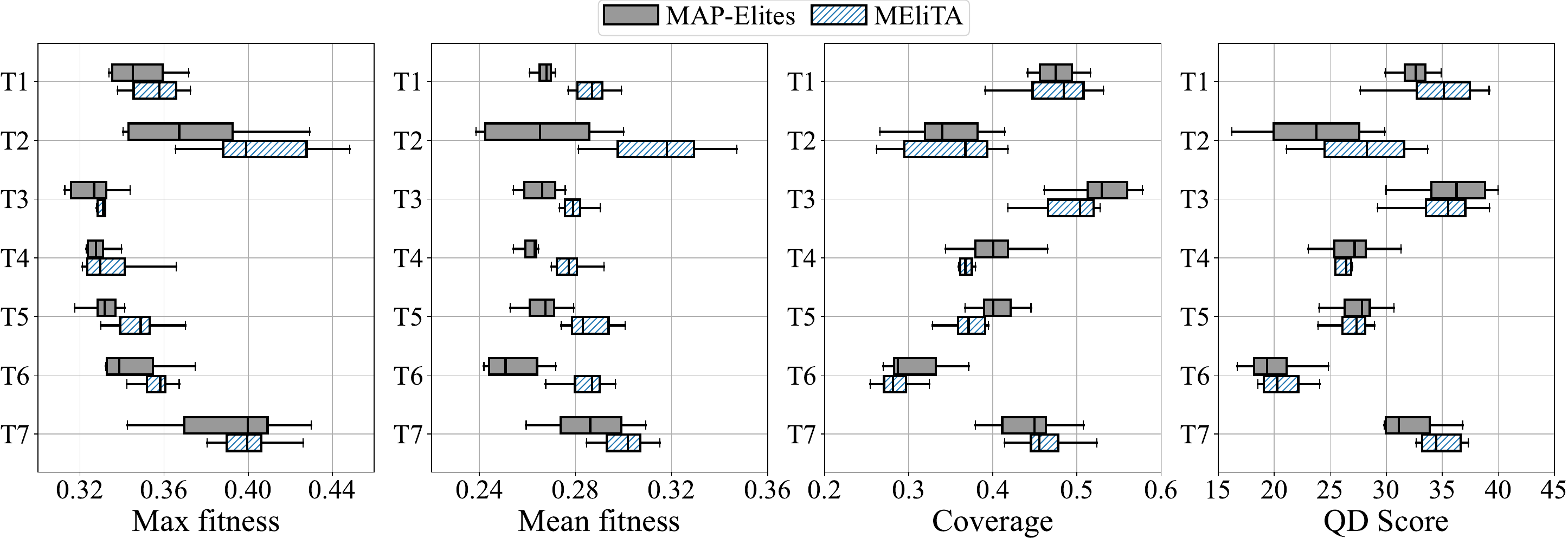}
    \caption{Metrics of the archives after 2000 selections in MAP-Elites and MEliTA. Box plots summarise values from 10 runs per title.}
    \label{fig:h1_plots}
\end{figure}

Figure \ref{fig:h1_plots} shows the four performance metrics of the QD algorithm applied on the final archive (after 2000 parent selections). It is evident that overall MEliTA results in fitter individuals than MAP-Elites, with results from 6 of 7 game titles having a statistically higher mean fitness (all except T7) and 3 of 7 game titles having a statistically higher maximum fitness (T2, T5, T6). This indicates that the evolved solutions are overall more coherent between the visuals and the text, likely due to the fact that a generated text or image can be paired with another image or text from the archive that is a better fit than the image-text combination produced by MAP-Elites. This prioritisation of pairing artefacts from different elites together leads to a lower coverage of the feature space. While for many game titles this drop in coverage is slight, MAP-Elites has significantly higher coverage for 2 of 7 game titles (T3 and T5). Since MEliTA produces fewer but fitter elites compared to MAP-Elites, QD scores of the two methods tend to be comparable. The only significant difference in QD score is for T7 where MEliTA has a higher QD score than MAP-Elites.
Based on this analysis, we can claim that MEliTA leads to better results at the cost, at times, of feature map coverage. Therefore, H1 is only partially validated---for quality but not diversity.

\subsection{Evaluating quality and diversity throughout evolution}\label{sec:results_h2}
\begin{figure}[t]
    \centering
    \includegraphics[width=0.98\textwidth]{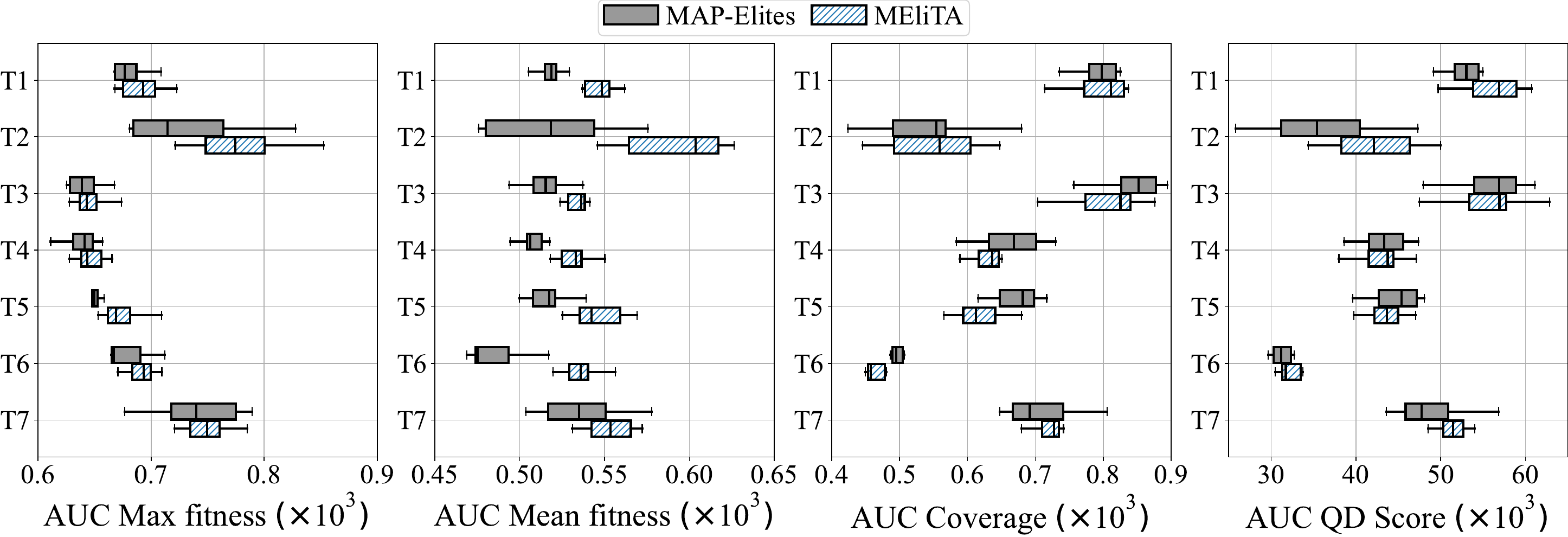}
    \caption{Area under curve (AUC) of QD metrics over 2000 selections in MAP-Elites and MEliTA. Box plots summarise values from 10 runs per title.}
    \label{fig:h2_plots}
\end{figure}
Figure \ref{fig:h2_plots} shows the area under the curve (AUC) scores over 2000 parent selections for the different QD performance metrics. The findings from the AUC corroborate those of Section \ref{sec:results_h1}: mean and maximum fitness rises faster in MEliTA than in MAP-Elites (significantly so in 3 and 4 out of 7 game titles respectively). Coverage rises faster in MAP-Elites than MEliTA in most cases (significantly so in 2 out of 7 game titles), and AUC of the QD score is mostly comparable between the two methods; MEliTA has significantly higher AUC for the QD score only for T1. Based on these findings, we can claim that MEliTA can find fitter individuals quicker than MAP-Elites without Transverse Assessment. H2 is thus only partially validated---for quality but not diversity.

\subsection{Evaluating visual and textual diversity of final artefacts}\label{sec:results_h3}
\begin{figure}[t]
    \centering
    \includegraphics[width=0.98\textwidth]{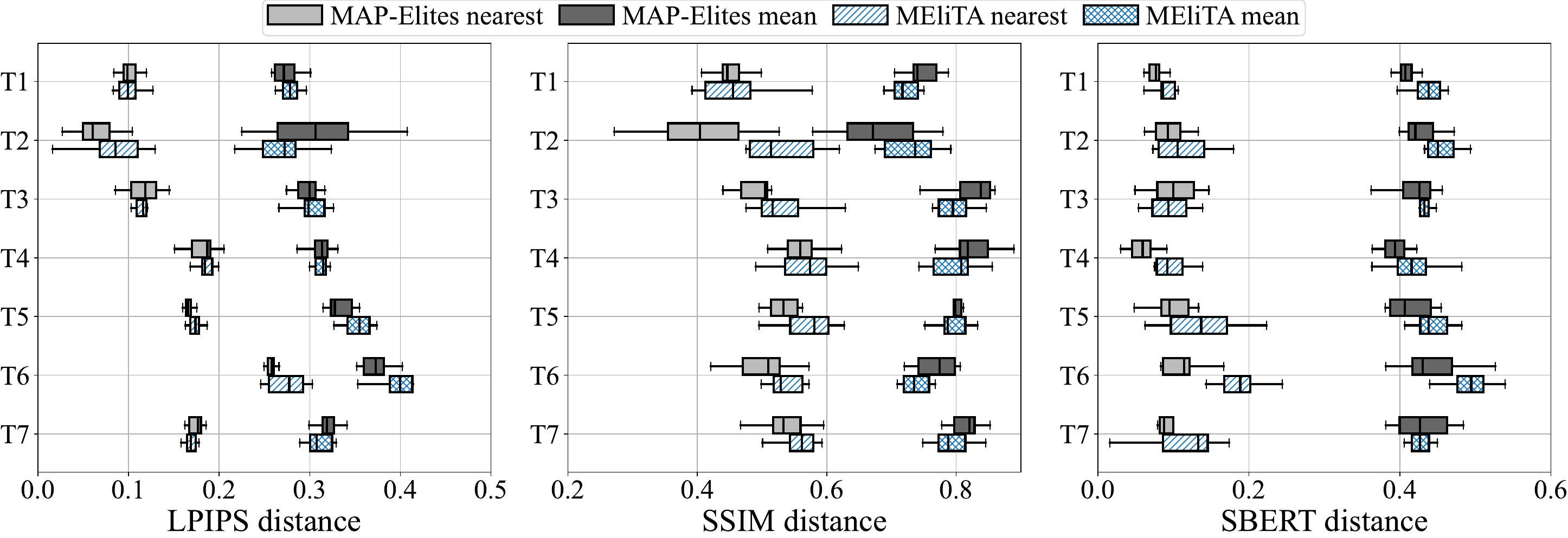}
    \caption{Visual and textual distance metrics (mean and nearest-neighbour) among final elites of MEliTA and MAP-Elites without Transverse Assessment. Box plots summarise values from 10 runs per title.}
    \label{fig:h3_plots}
\end{figure}
To assess the perceivable differences in the two methods' archives, we use visual distances (via LPIPS and SSIM) and textual distance (via SBERT embeddings) of the final elites in each run of MEliTA and MAP-Elites after 2000 parent selections. Figure \ref{fig:h3_plots} shows the mean and nearest-neighbour distances for each of these metrics, which are largely orthogonal to the BCs used in MAP-Elites. 

Figure \ref{fig:h3_plots} indicates that, as a whole, the final results of MAP-Elites are not more or less diverse than MEliTA. In terms of visuals, MEliTA has significantly higher mean and nearest-neighbour LPIPS distance than MAP-Elites in results of 2 game titles (T5, T6) and 1 game title (T5) respectively, and significantly higher nearest-neighbour SSIM distance than MAP-Elites in 2 game titles (T2, T3). While these findings are hardly consistent, they indicate that MEliTA tends to produce more diverse images than MAP-Elites. Clearer patterns are gleaned for textual diversity: MEliTA has significantly higher mean and nearest-neighbour SBERT distances in results of 3 game titles (T1, T2, T6) and 4 game titles (T1, T4, T5, T6) respectively. No game title has significant diversity improvements for MEliTA across all metrics. However, we can claim that H3 is validated: the final archives of MEliTA, even if smaller, are more visually and textually diverse than MAP-Elites without Transverse Assessment. 

\newcommand{\samplewidth}{0.19\textwidth}
\begin{table}[t]
    \centering
    \caption{Example output of MEliTA for the fictitious game title ``The Shadow Warrior 2: Shadows of the Past'' (T6). The description (after removal of the title) is left with its original errors.}
    \label{tab:t5_clusters}
    \begin{tabular}{|p{\samplewidth}|p{\samplewidth}|p{\samplewidth}|p{\samplewidth}|p{\samplewidth}|}
    \hline
    \multicolumn{1}{|c|}{Game 1} & \multicolumn{1}{c|}{Game 2} & \multicolumn{1}{c|}{Game 3} & \multicolumn{1}{c|}{Game 4} & \multicolumn{1}{c|}{Game 5} \\ 
    \includegraphics[width=\samplewidth]{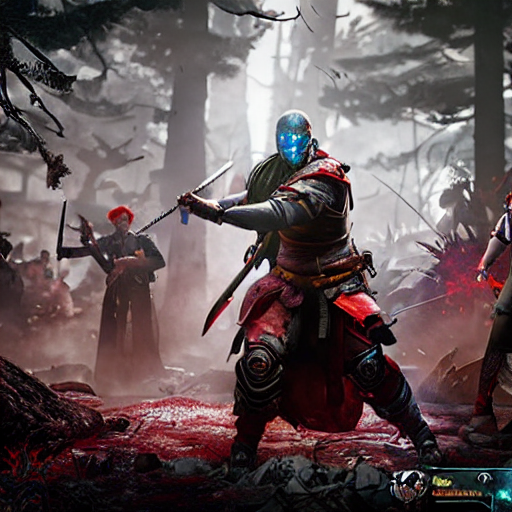} &  
    \includegraphics[width=\samplewidth]{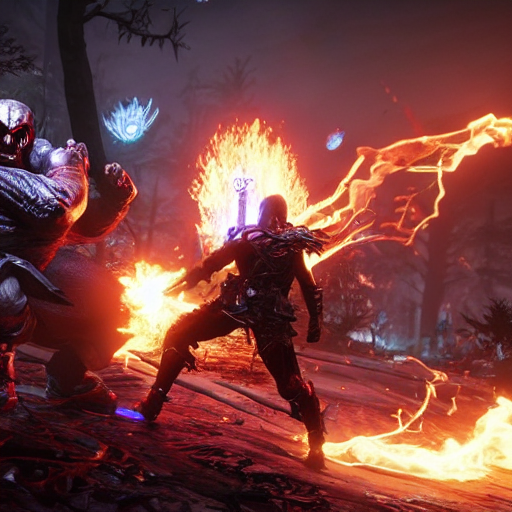} & 
    \includegraphics[width=\samplewidth]{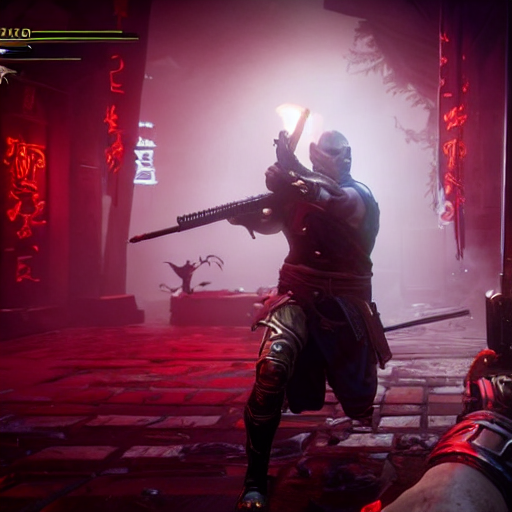} & 
    \includegraphics[width=\samplewidth]{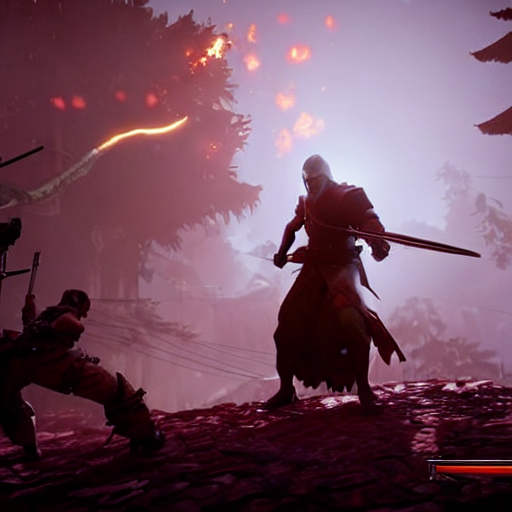} & 
    \includegraphics[width=\samplewidth]{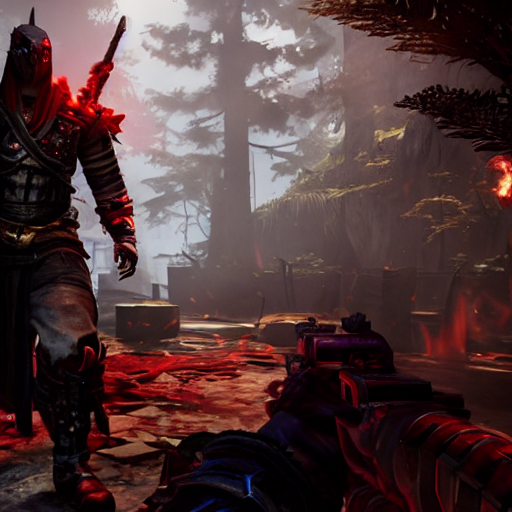} \\
\raggedright\tiny{In this game, players have to use their sword and weapons wisely because there are many enemies who will try very hard for you. After many defeats in a long time, each one offers new challenges.}
&
\raggedright\tiny{After a global catastrophe, you are left alone to face your past. The greatest evil - an alien race called Korda – is planning on tearing Earth apart as they do every few centuries in this new adventure from Arkane Studios and Infamous Games.}
&
\raggedright\tiny{Play as a shadow warrior, taking on different threats in this third person shooter that's more brutal than ever before.}
&
\raggedright\tiny{In this game, players have to use their swords and bows in order not only survive but also fight with others.}
&
\raggedright\tiny{After years' long struggle, Dark Lord Arthur's dark legacy is coming to light, and a new evil seems lurking at every corner. TheShadow Warrior2 : Shadows OfthePast.}
\tabularnewline
\hline
\end{tabular}
\end{table}

It is also interesting to observe differences in elites directly. We choose T6 (``The Shadow Warrior 2: Shadows of the Past'') as it has high LPIPS and SBERT distances overall, and an indicative run that had high values for both. We apply $k$-medoid clustering using a Euclidean distance combining SBERT and LPIPS. The medoids for $k=5$ are shown in Table~\ref{tab:t5_clusters}.
These samples' text descriptions vary, with some merely describing the gameplay (Games 1, 3 and 4) and some describing a science fiction narrative (Game 2) or a Camelot narrative (Game 5). Some text descriptions include ``noise'' (e.g. the end of the description in Game 5) while others overlap due to the partial mutation operation (e.g. Game 1 and Game 4). In terms of the images, most depict ``action shots'' which are likely not part of gameplay, although Game 5 does show a first-person shooter view (presumably in-game) complete with a gun within an otherwise fantasy setting. Depicted characters are mostly armoured and their faces covered, although it is unclear whether this is because of ninja tropes or because of the negative prompts on e.g. ``bad anatomy''. Characters are often seen wielding swords, although there are also modern guns shown in Games 3 and 5. While backgrounds mostly show a foggy forest setting, Game 2 deviates with warmer colours and fiery ``wings'' on the character at the image centre. Overall, while the images of these medoids do not depict as much diversity as one would expect (in terms of ``ninja'' characters and colour palettes), there is an overall consistency between the (sometimes generic) descriptions and the associated images. Perhaps the most concerning fact is that Game 2 includes names of actual game studios, likely due to the Steam dataset that GPT-2 was trained on; such text could be problematic due to intellectual property concerns if the use case would be made widely available.

\section{Discussion}
\label{sec:discussion}

Through the use case for generating fictional game descriptions and cover images, we ascertained that MEliTA can produce fitter---if fewer---elites. We also recognise that coverage decreases since new elites are only added to the archive if there is no better alternative when pairing an offspring's changed modality with the modalities of an existing elite. This drop in coverage did not result in a drop in visual and textual diversity of produced elites when assessed on metrics decoupled from the BCs. Both in terms of mean distance and nearest-neighbour distance (which can counter the impact of a different number of elites), the elites of MEliTA were more visually and semantically diverse, as well as more coherent (higher fitness) than their counterparts produced without transverse assessment. The diversity metrics (SBERT, LPIPS, SSIM) represent the state-of-the-art for these purposes, but should be corroborated with human feedback. Future work could explore, via a user study with players and game developers, to which degree the generated multimodal artefacts are deemed diverse or inspirational.

It is worth noting that MEliTA produces far more candidate solutions (via transverse assessment) than the single offspring produced by MAP-Elites. This means that for MEliTA the number of fitness evaluations can be much higher than for MAP-Elites, although the number of BC evaluations is the same (only one individual with the parent's unchanged modality). For a $16\times16$ feature map, as in this use case, MEliTA may perform as many as 16 fitness evaluations per parent selection compared to 1 fitness evaluation for MAP-Elites. Calculating the CLIP score is not expensive, so the computational overhead of MEliTA is negligible; this may not be the case for simulation-based fitness evaluation \cite{mouret2015illuminating}.

The use case in this paper included a crude form of constrained optimisation \cite{coello2010survey}, as the death penalty was applied on all unclassified text descriptions \cite{michalewicz1995dontkill}. Additions to MEliTA could explore better ways of handling infeasible individuals, e.g. via a two-population approach as in \cite{sfikas2022general,khalifa2018talakat}. Preliminary experiments using additional cells for unclassified individuals did not yield any substantial differences from MEliTA, but future work could combine constrained QD \cite{sfikas2022general} with a minimal fitness threshold to distinguish feasible individuals.

Examples of the generated results in Table \ref{tab:t5_clusters} also highlighted some limitations in the chosen variation operators. By operating on the phenotype (images or text) rather than on a latent representation, variation operators produce more controllable and less noisy output---compared to latent variable evolution \cite{fontaine2021differentiable,zammit2022seeding}. Text variation specifically tends to cause overlap between individuals' descriptions due to partial mutation (see Section \ref{sec:usecase_text_generation}). Alternative text mutation operators could leverage models capable of generating text tokens in reverse \cite{west2021reflective}, i.e. generating the first part of the descriptions given the end part, or use genotypic operators which alter the embeddings of the text \cite{lehman2023evolllms}. Moreover, the GPT-2 trained model is admittedly dated in the current ecosystem of LLMs. While future work could explore a more modern text generator such as OpenAI's ChatGPT \cite{openai2023gpt4}, the main challenge is its closed-source nature which hinders both fine-tuning its weights and keeping track of the provenance of its output.

While this use case explored a bimodal problem that was easy to visualise in a feature map with two axes, MEliTA could conceivably work with more artefact modalities and more BC dimensions---and those two do not need to necessarily match. Each artefact modality could easily have more than one BC; even in our case the two metrics for images (complexity and colourfulness) could have become different BC dimensions leading to a 3-dimensional feature map. Moreover, MEliTA could be applied to a more complex multi-modal creative problem such as evolving text, image, and audio for an interactive story application. In that case, an LLM can be used to generate the story and the locations over which it takes place, SD can be used to generate background images per location, and a text-to-music generator such as MusicGen \cite{copet2023simple} can generate a soundscape for each part of the story. In this case fitness can be assessed via a multimodal network such as ImageBind \cite{girdhar2023imagebind}, by computing the vector similarity of each embedding.

\section{Conclusion}
\label{sec:conclusion}

This paper aimed to address the recent emergence of image-to-text and text-to-image generators through a QD perspective. To this end, we adapt MAP-Elites to operate on solutions that consist of artefacts of different modalities, and implement a transverse assessment method that allows such (partial) artefacts to be shared with other individuals. We show that MEliTA can outperform MAP-Elites, at the cost of fewer solutions. 
Extensions of this work should explore more state-of-the-art (but open-source) text generation algorithms while also extending the transverse assessment approach to incorporate quality constraints and integrate more types of artefacts and modalities.

\section{Acknowledgements}
This project has received funding from the European Union’s Horizon 2020 programme under grant agreement No 951911.

\bibliographystyle{splncs04}

\end{document}